# Self-Motions of General 3-<u>R</u>PR Planar Parallel Robots

**Sébastien Briot[*], Ilian A. Bonev[**], Damien Chablat[***],**

**Philippe Wenger[***] and Vigen Arakelian[*]**

[*] *Département de Génie Mécanique et Automatique,*

*Institut National des Sciences Appliquées (INSA), Rennes, France*

*(sebastien.briot@ens.insa-rennes.fr, vigen.arakelyan@insa-rennes.fr)*

[**] *Department of Automated Production Engineering,*

*École de Technologie Supérieure (ÉTS), Montréal, Canada*

*(ilian.bonev@etsmtl.ca)*

[***] *Institut de Recherche en Communications et Cybernétique de Nantes*

*1, rue la Noë – BP 92101 44321, Nantes Cedex 3, France*

*(damien.chablat@irccyn.ec-nantes.fr, philippe.wenger@irccyn.ec-nantes.fr)*

ABSTRACT

This paper studies the kinematic geometry of general 3-<u>R</u>PR planar parallel robots with actuated base joints. These robots, while largely overlooked, have simple direct kinematics and large singularity-free workspace. Furthermore, their kinematic geometry is the same as that of a newly developed parallel robot with SCARA-type motions. Starting from the direct and inverse kinematic model, the expressions for the singularity loci of 3-<u>R</u>PR planar parallel robots are determined. Then, the global behaviour at all singularities is geometrically described by studying the degeneracy of the direct kinematic model. Special cases of self-motions are then examined and the degree of freedom gained in such special configurations is kinematically interpreted. Finally, a practical example is discussed and experimental validations performed on an actual robot prototype are presented.

## 1. Introduction

From an industrial point of view, the complexity and the existence of numerous singular configurations seem to be the worse drawback of parallel robots because these configurations reduce the size of the workspace, which is already smaller than that of similarly-sized serial robots. Fortunately, the determination of singularities is a well studied problem and several computational methods have already been presented (Gosselin 1990; Zlatanov et al. 1994; Bonev et al. 2003).

The worst singular configuration a parallel robot can meet is the Type 2 singularity (Gosselin 1990). In such a singularity, the robot gains at least one degree of freedom and cannot resist some wrenches applied to its platform. Furthermore, the robot cannot exit such a singular configuration, without external help. Type 2 singular configurations can be divided into two classes, depending on the nature of the degree(s) of freedom gained, being either infinitesimal or finite, i.e., *self-motion*. However, merely studying the Jacobian (Gosselin 1990; Bonev et al. 2003), one cannot identify the nature of Type 2 singularities.

Symmetry and, more precisely, design conditions that simplify the generally too complex direct kinematics of parallel robots are often privileged by robot designers. Unfortunately, such design conditions usually lead to self-motions, which are certainly the worst type of singularity. Furthermore, as we show in this paper, self-motions also occur in unsymmetrical seemingly general designs without simplified direct kinematic models. Hence, it is essential that the design conditions for such self-motions be well known in order to be avoided.

Several papers discuss self-motions in parallel robots. Not surprisingly, most of them deal with the Gough-Stewart platform, whose direct kinematic model leads to as much as forty real solutions, for a relatively general design. Design conditions simplifying the direct kinematics of Gough-Stewart platforms, and subsequently leading to self-motions, are given in (Karger 2001; Karger 2003; Karger and Husty 1998; Husty and Karger 2000; Husty and Zsombor-Murray 1994; Wohlhart 2003). A classification of all self-motions of the Stewart-Gough platform is presented in (Karger and Husty 1998). It is shown that the self-motions can be translations, pure rotations, generalized screw motions, motions equivalent to the displacements of spherical four-bar mechanisms, or more complex spatial motions.

The Stewart-Gough platform is not the only parallel robot with self-motions. A few other parallel robots having self-motions have also been studied. For example, in (Bonev et al. 2006), it is shown that all singularities of the special 3-$\underline{R}RR$ ($R$ stands for a passive revolute joint, and $\underline{R}$ for an actuated revolute joint) spherical parallel robot, known as the Agile Eye, are self-motions. The analysis of self mobility of spatial 5$R$ closed-loop mechanisms with one degree of freedom are presented in (Karger 1998). Reference (Bandyopadhyay and Ghosal 2004) discusses the determination of generalized analytical expressions for the analysis of self-motions and presents several examples for both planar and spatial mechanisms with legs composed of $R$ joints.

In this paper, we will study the self-motions of general 3-$\underline{R}PR$ planar parallel robots ($P$ stands for a passive prismatic joint). The 3-$\underline{R}PR$ planar parallel robot has a simple direct kinematic model and, when properly designed, a relatively large singularity-free workspace. However, despite these advantages, only a couple of works deal with this kind of robot (Hayes 1999; Hayes and Zsombor-Murray 2004). Yet, a recently developed new decoupled parallel robot with SCARA-type motions (Briot and Arakelian 2007) has its planar displacements governed by the same kinematic model as that of a 3-*<u>R</u>PR* planar parallel robot. Furthermore, the self-motions of a particular design of a 3-*<u>R</u>PR* planar parallel robot with congruent equilateral base and platform were studied in (Chablat et al. 2006), mainly from a theoretical point of view. This paper basically generalizes this study and demonstrates the advantages of general 3-*<u>R</u>PR* planar parallel robots.

The rest of this paper is organised as follows. Section 2 deals with the kinematics of the general 3-*<u>R</u>PR* planar parallel robot. The direct and inverse kinematic models are derived from the closure equations, and the singularity analysis based on the observation of the rank of the Jacobian matrix is presented. Section 3 presents a self-motion analysis based on the degeneracy of the direct kinematic model. Singularity loci are given and the degree of freedom gained is kinematically interpreted. Section 4 deals with a particular case of 3-*<u>R</u>PR* planar parallel robot with equilateral base and platform triangles and the results obtained are validated on an actual robot prototype. Conclusions are given in Section 5.

## 2. Kinematics and singularity analysis

The following analysis is based on the schematics of the robot shown in Fig. 1. The revolute joints $A_i$ (in the remainder of this paper, $i$ = 1, 2, 3) are fixed on the base and are actuated. Each leg is composed of one passive prismatic joint, placed between points $A_i$ and $B_i$, and one passive revolute joint $C_i$, connected to the mobile platform.

We consider that we control the position ($x$, $y$) of point $P$ from the mobile platform and the orientation $\phi$ of the mobile platform. The origin of the base frame is chosen at point $O$. Points $O$ and $P$ are located at the centres of the circumscribed circles of triangles $A_1A_2A_3$ and $C_1C_2C_3$, respectively (Fig. 2). Finally, let $\rho_i = |A_iB_i|$ and $L_i = |B_iC_i|$, the latter, referred to as an *offset*.

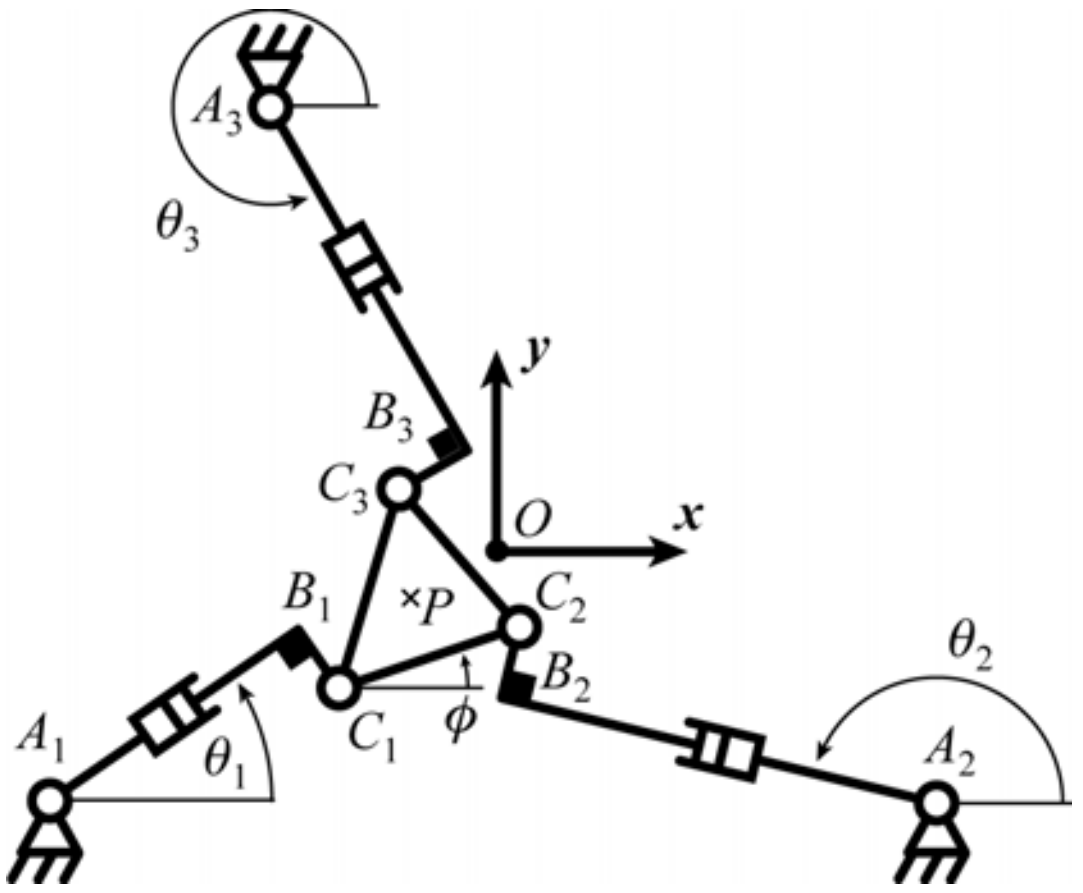

**Figure 1. Schematic representation of the 3-*__R__PR* planar parallel robot under study.**

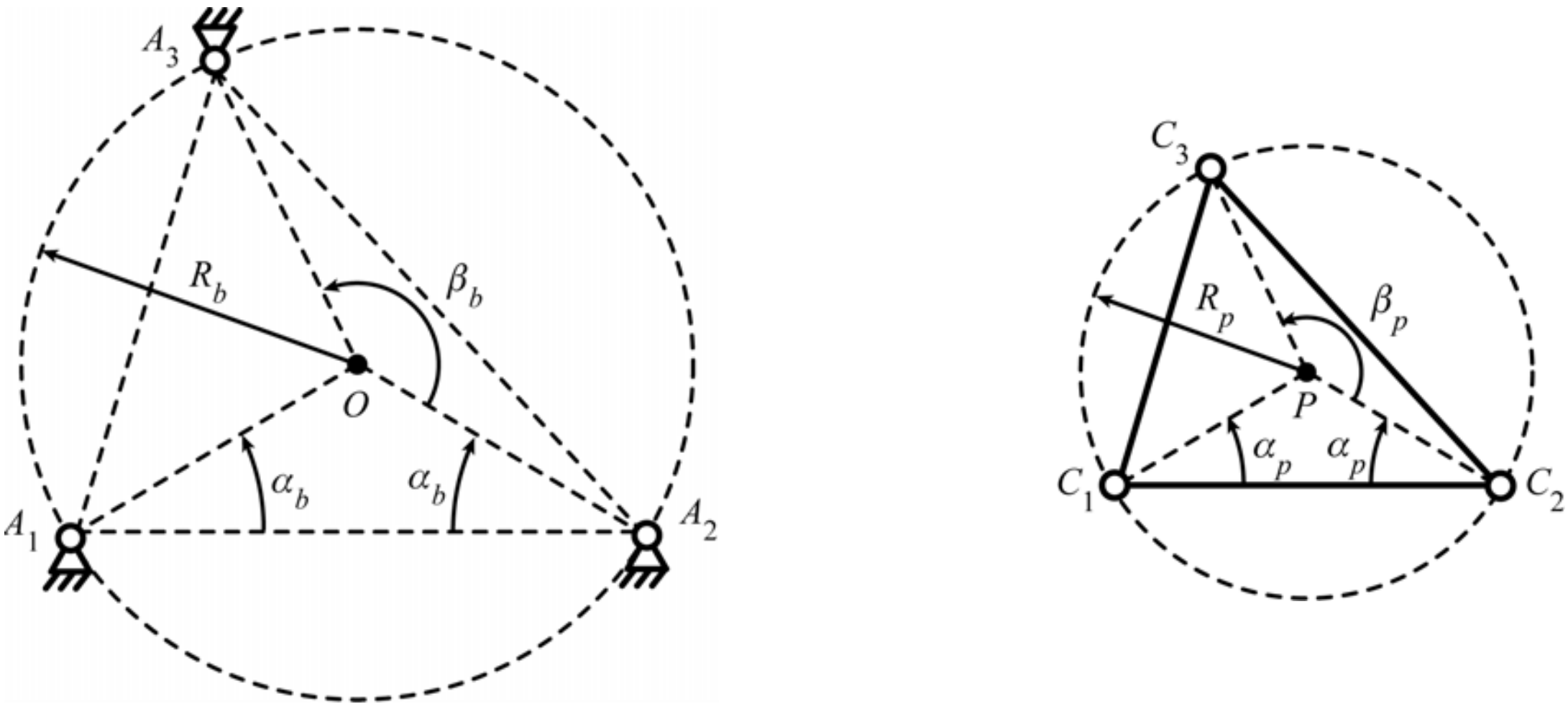

**(a) fixed base** **(b) mobile platform**

**Figure 2. Parameterisation of the base and platform triangles.**

Thus, it is possible to express the position of points $A_i$ and $C_i$ as

$$\mathbf{OA}_i = \begin{bmatrix} x_{Ai} \\ y_{Ai} \end{bmatrix} = R_b \begin{bmatrix} \cos\gamma_i \\ \sin\gamma_i \end{bmatrix}, \ \mathbf{OC}_i = \begin{bmatrix} x_{Ci} \\ y_{Ci} \end{bmatrix} = \begin{bmatrix} x \\ y \end{bmatrix} + R_p \begin{bmatrix} \cos(\phi+\delta_i) \\ \sin(\phi+\delta_i) \end{bmatrix}, \tag{1}$$

where $\gamma_i = (\alpha_b + \pi, -\alpha_b, -\alpha_b + \beta_b)$ and $\delta_i = (\alpha_p + \pi, -\alpha_p, -\alpha_p + \beta_p)$. From these expressions and referring to (Bonev et al. 2003), one can determine the closure equations of the system:

$$\mathbf{OC}_i - \mathbf{OB}_i = \begin{bmatrix} x_{Ci} - x_{Ai} - \rho_i \cos\theta_i \\ y_{Ci} - y_{Ai} - \rho_i \sin\theta_i \end{bmatrix} = L_i \begin{bmatrix} -\sin\theta_i \\ \cos\theta_i \end{bmatrix}. \tag{2}$$

Skipping the derivation and referring the reader to (Bonev et al. 2003), the velocity equation for the 3-*<u>R</u>PR* robot is:

$$\mathbf{A}\left[\dot{\phi}, \dot{x}, \dot{y}\right]^T = \mathbf{B}\left[\dot{\theta}_1, \dot{\theta}_2, \dot{\theta}_3\right]^T \tag{3}$$

with

$$\mathbf{A} = \begin{bmatrix} \mathbf{f}_1^T \mathbf{E} \mathbf{g}_1 & \mathbf{f}_1^T \\ \mathbf{f}_2^T \mathbf{E} \mathbf{g}_2 & \mathbf{f}_2^T \\ \mathbf{f}_3^T \mathbf{E} \mathbf{g}_3 & \mathbf{f}_3^T \end{bmatrix}, \; \mathbf{B} = \begin{bmatrix} \rho_1 & 0 & 0 \\ 0 & \rho_2 & 0 \\ 0 & 0 & \rho_3 \end{bmatrix}, \tag{4}$$

and

$$\mathbf{g}_i = \left[x_{Ci} - x \quad y_{Ci} - y\right]^T, \; \mathbf{E} = \begin{bmatrix} 0 & 1 \\ -1 & 0 \end{bmatrix}, \; \mathbf{f}_i = \begin{bmatrix} -\sin\theta_i \\ \cos\theta_i \end{bmatrix}. \tag{5}$$

## 2.1. Inverse kinematic problem

Solving the inverse kinematics for each leg of this robot is essentially finding the intersection points between two circles, one with diameter $|A_iC_i|$ centred at the middle of segment $A_iC_i$, and one with radius $L_i$ and centred at $C_i$. Premultiplying both sides of equation (2) with the term $\mathbf{f}_i^T$, one can obtain an equation expressing the angles $\theta_i$ as function of the other parameters:

$$(x_{Ci} - x_{Ai})\sin\theta_i - (y_{Ci} - y_{Ai})\cos\theta_i - L_i = 0. \tag{6}$$

From equation (6), it is possible to find the expressions for the active-joint variables $\theta_i$ as functions of the position $(x, y)$ and the orientation $\phi$ of the mobile platform:

$$\theta_{ip} = 2\tan^{-1}\left(\frac{-(x_{Ci}-x_{Ai})+\sqrt{(x_{Ci}-x_{Ai})^2+(y_{Ci}-y_{Ai})^2-L_i^2}}{-L_i+y_{Ci}-y_{Ai}}\right), \tag{7a}$$

$$\theta_{im} = 2\tan^{-1}\left(\frac{-(x_{Ci}-x_{Ai})-\sqrt{(x_{Ci}-x_{Ai})^2+(y_{Ci}-y_{Ai})^2-L_i^2}}{-L_i+y_{Ci}-y_{Ai}}\right). \tag{7b}$$

The two solutions $\theta_{ip}$ and $\theta_{im}$ define the two inverse kinematic solutions for leg $i$ (Fig. 3). These define a total of eight solutions to the inverse kinematics of the parallel robot, also called *working modes* (Wenger and Chablat 1998). We will see that for this robot, and provided nonzero offsets, $L_i > 0$, the singularity loci will depend on the working mode.

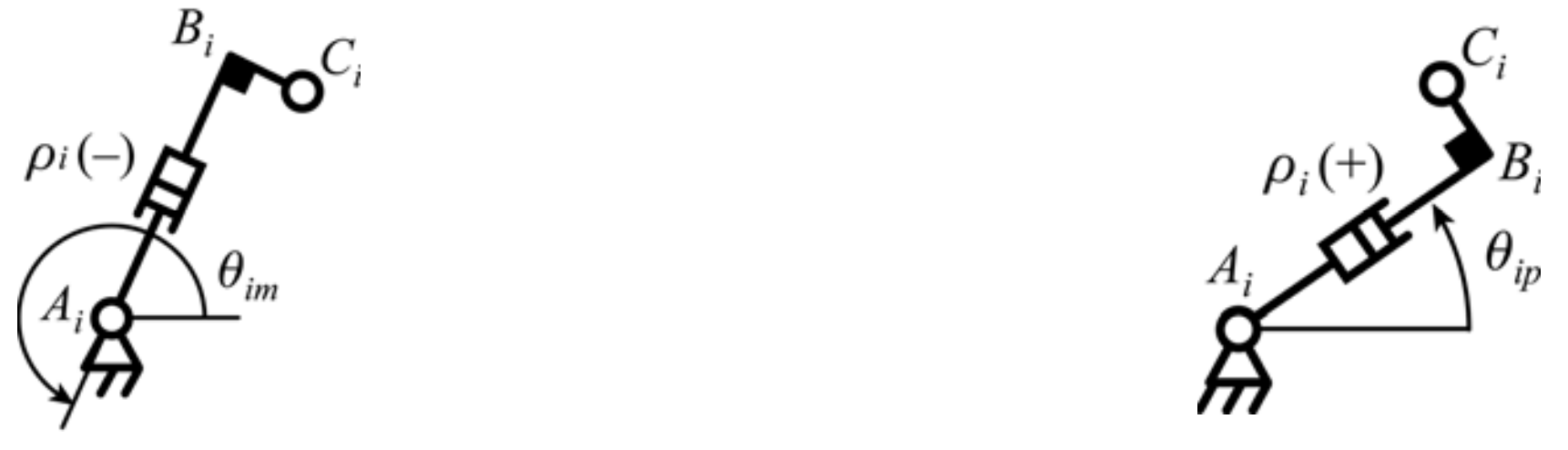

**(a) first solution, $\rho_i$ (+)** **(b) second solution, $\rho_i$ (−)**

**Figure 3. The two inverse kinematic solutions of the $i^{\text{th}}$ leg of the robot.**

2.2. Type 1 singularities

Type 1 singularities occur when the determinant of **B** vanishes, i.e., when $\rho_i = 0$ (for $i = 1, 2$, or 3) (Fig. 4) (Bonev et al. 2003). These configurations correspond to the internal boundaries of the workspace of a general 3-*<u>R</u>PR* planar parallel robot. When the offsets are zero, $L_i = 0$, there is a generic Type 1 (RI) singularity where the input velocities are indeterminate (Zlatanov et al. 1994). On this singularity, the inverse kinematic model of leg $i$ admits only one solution because $(x_{Ci}-x_{Ai})^2+(y_{Ci}-y_{Ai})^2-L_i^2=\rho_i^2=0$.

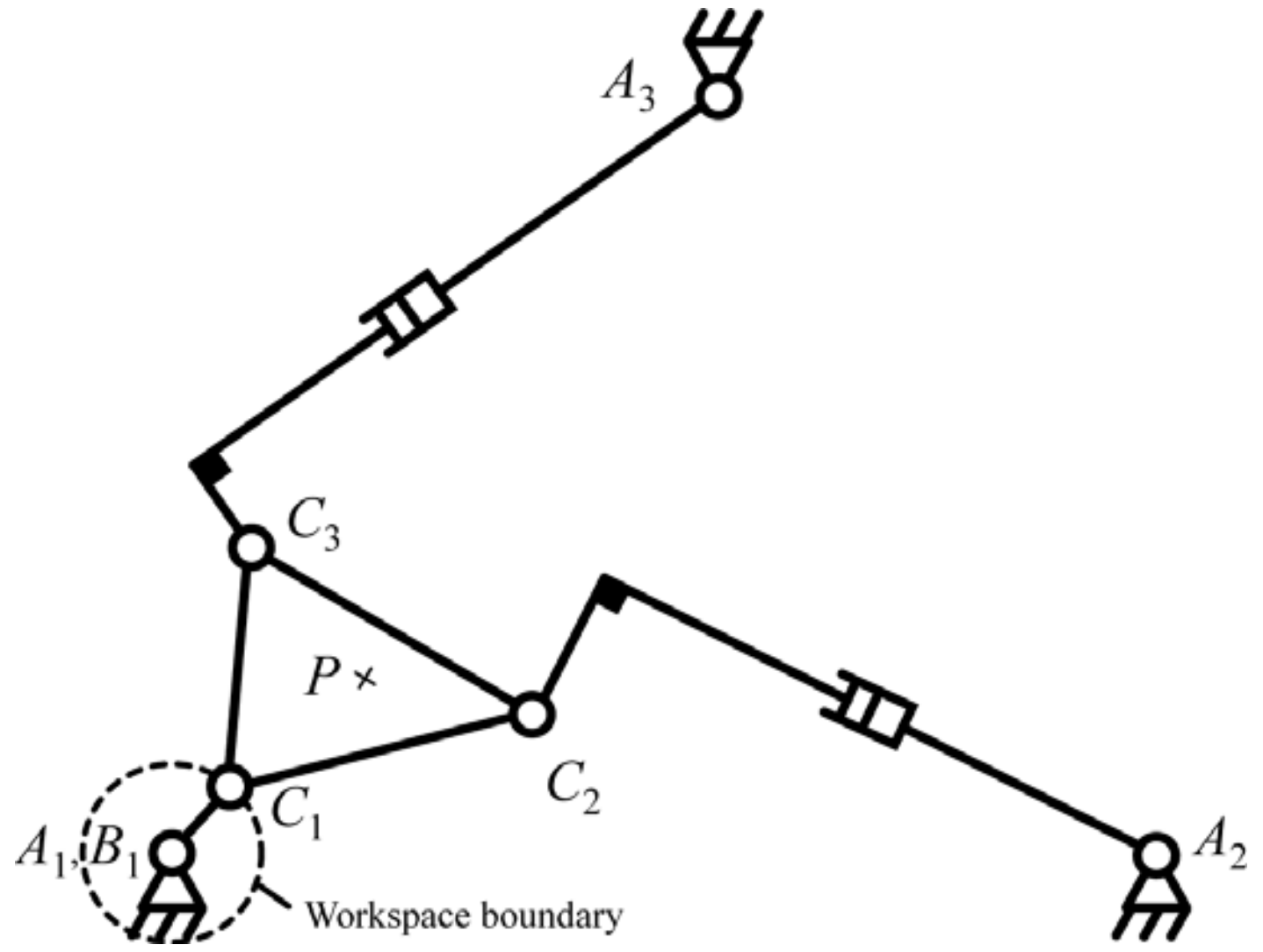

**Figure 4. Type 1 singularity.**

2.3. Direct kinematic problem

It is shown in (Merlet 1996) that the solution of the direct kinematics of a 3-$\underline{R}PR$ planar parallel robot is equivalent to finding the intersection points between an ellipse and a line, but no analytical expressions are given. Let us dismount the revolute joint at $C_3$. For given active-joint variables $\theta_1$ and $\theta_2$, points $C_1$ and $C_2$ are constrained to move along two lines, $\mathcal{L}_1$ and $\mathcal{L}_2$, respectively, and the mobile platform undergoes a Cardanic motion (Sekulie 1998; Tischler et al. 1998) (Fig. 5). As a result, any points $Q$ from the mobile platform, including $P$ and $C_i$, describe a curve $\mathcal{E}(Q)$, which can be an ellipse, two parallel lines or a doubly-traced line segment. Thus, the direct kinematics can be solved by finding the intersection points between the curve $\mathcal{E}(C_3)$ and the line $\mathcal{L}_3$.

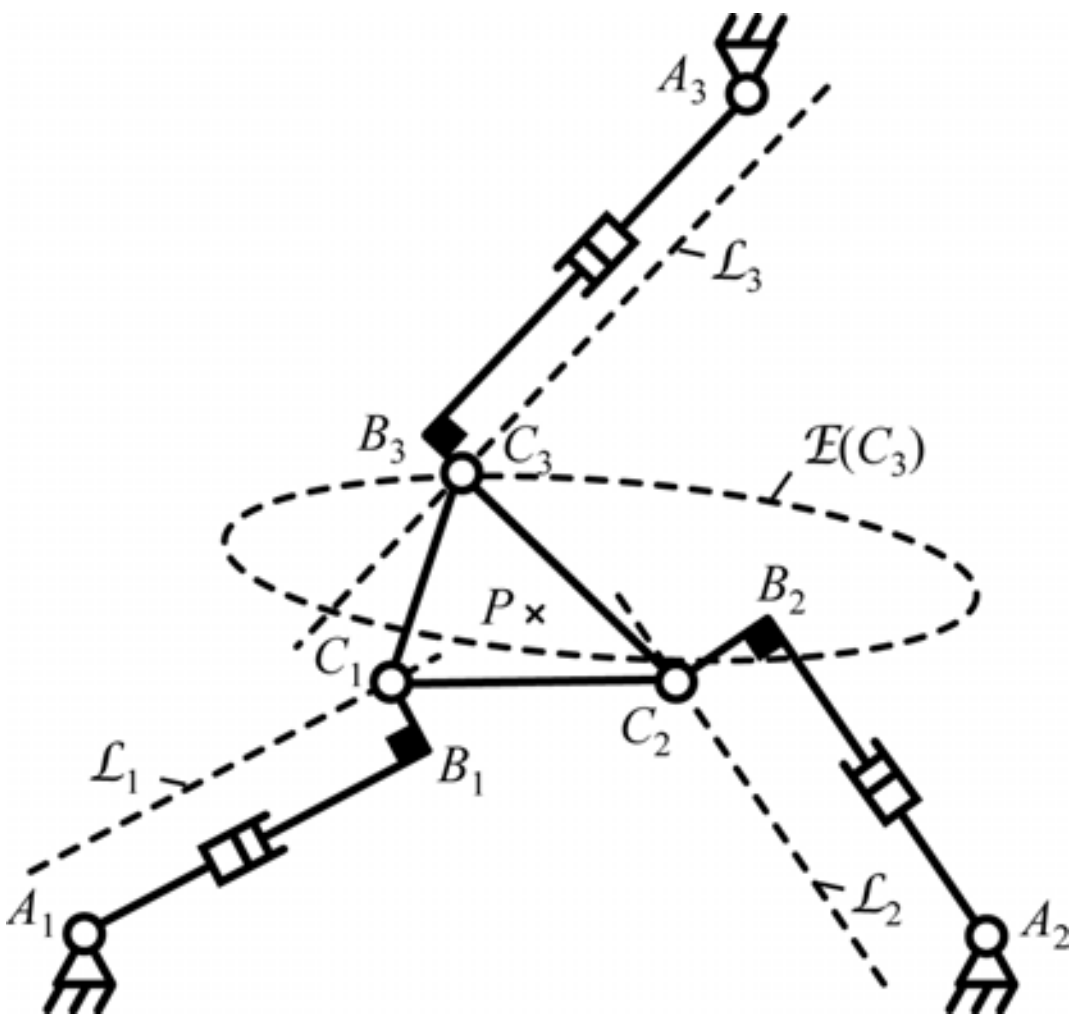

**Figure 5. Geometric interpretation of the direct kinematics.**

Let us now derive the expression of the elliptic curve $\mathcal{E}(C_3)$. It is possible to write the following closure equation:

$$\mathbf{OC_3} = \mathbf{OA_1} + \mathbf{A_1B_1} + \mathbf{B_1C_1} + \mathbf{C_1C_3}. \tag{8}$$

This yields the following expression:

$$\mathbf{OC_3} = \begin{bmatrix} x_{C3} \\ y_{C3} \end{bmatrix} = \begin{bmatrix} x_{A1} \\ y_{A1} \end{bmatrix} + \rho_1 \begin{bmatrix} \cos\theta_1 \\ \sin\theta_1 \end{bmatrix} + L_1 \begin{bmatrix} -\sin\theta_1 \\ \cos\theta_1 \end{bmatrix} + 2R_p \cos\left(\frac{\beta_p}{2} - \alpha_p\right) \begin{bmatrix} \cos\left(\frac{\beta_p}{2} + \phi\right) \\ \sin\left(\frac{\beta_p}{2} + \phi\right) \end{bmatrix}. \tag{9}$$

In this expression, all parameters are known except $\rho_1$ and $\phi$. However, they are dependent on each other. Without loss of generality, we choose $\phi$ as independent variable and express $\rho_1$ as a function of $\phi$, using the following closure equation:

$$\mathbf{A_1A_2} = \mathbf{A_1B_1} + \mathbf{B_1C_1} + \mathbf{C_1C_2} + \mathbf{C_2B_2} + \mathbf{B_2A_2}. \tag{10}$$

Developing this relation, we obtain:

$$\begin{bmatrix} x_{A2} - x_{A1} \\ y_{A2} - y_{A1} \end{bmatrix} = \rho_1 \begin{bmatrix} \cos\theta_1 \\ \sin\theta_1 \end{bmatrix} + L_1 \begin{bmatrix} -\sin\theta_1 \\ \cos\theta_1 \end{bmatrix} + 2R_p \cos\alpha_p \begin{bmatrix} \cos\phi \\ \sin\phi \end{bmatrix} - L_2 \begin{bmatrix} -\sin\theta_2 \\ \cos\theta_2 \end{bmatrix} - \rho_2 \begin{bmatrix} \cos\theta_2 \\ \sin\theta_2 \end{bmatrix}. \tag{11}$$

Expressing $\rho_1$ and $\rho_2$ as a function of $\phi$ from equation (11), we obtain:

$$\rho_j = a_{j1} + a_{j2}\cos\phi + a_{j3}\sin\phi, \quad (j = 1, 2) \tag{12}$$

where the expressions for $a_{ji}$ are given in the appendix. Reintroducing equation (12) in equation (9), we find the following relation:

$$\mathbf{OC_3} = \begin{bmatrix} x_{C3} \\ y_{C3} \end{bmatrix} = \begin{bmatrix} b_{11} + b_{12}\cos\phi + b_{13}\sin\phi \\ b_{21} + b_{22}\cos\phi + b_{23}\sin\phi \end{bmatrix}, \tag{13}$$

where $b_{ji}$ ($j = 1, 2$) are given in the appendix.

Thus for any fixed input parameters $\theta_i$, we have found in equation (13) the parametric expression of the elliptic curve $\mathcal{E}(C_3)$ depending on the orientation $\phi$ of the platform. Furthermore, we know that point $C_3$ belongs to line $\mathcal{L}_3$, whose expression is:

$$y = \tan\theta_3(x + L_3\sin\theta_3 - x_{A3}) + y_{A3} + L_3\cos\theta_3 . \tag{14}$$

Thus, the intersections between $\mathcal{E}(C_3)$ and $\mathcal{L}_3$ can be found by substituting $x$ and $y$ in equation (14) by the expressions of $x_{C3}$ and $y_{C3}$ of equation (13). After the substitution in equation (14) and multiplying the equation by $\cos\theta_3$, we obtain:

$$0 = \sin\theta_3(x_{C3} + L_3\sin\theta_3 - x_{A3}) + \cos\theta_3(y_{A3} + L_3\cos\theta_3 - y_{C3}) . \tag{15}$$

Developing equation (15),

$$c_1 + c_2\cos\phi + c_3\sin\phi = 0, \tag{16}$$

where $c_i$ are given in the appendix. Thus, from (16), it is possible to find the solution for $\phi$:

$$\phi = 2\tan^{-1}\left(\frac{-c_3 \pm \sqrt{c_3^2 - c_1^2 + c_2^2}}{c_1 - c_2}\right). \tag{17}$$

Note that this solution is not unique and corresponds to the two assembly modes of the robot. Finally, it is possible to find the expression for the position using the following closure equation:

$$\mathbf{OP} = \mathbf{OA_1} + \mathbf{A_1B_1} + \mathbf{B_1C_1} + \mathbf{C_1P}\ , \tag{18}$$

which yields:

$$\mathbf{OP} = \begin{bmatrix} x \\ y \end{bmatrix} = \begin{bmatrix} x_{A1} \\ y_{A1} \end{bmatrix} + \rho_1 \begin{bmatrix} \cos\theta_1 \\ \sin\theta_1 \end{bmatrix} + L_1 \begin{bmatrix} -\sin\theta_1 \\ \cos\theta_1 \end{bmatrix} + R_p \begin{bmatrix} \cos(\phi+\alpha_p) \\ \sin(\phi+\alpha_p) \end{bmatrix}. \tag{19}$$

2.4. Type 2 singularities analysis

Type 2 singularities occur when the determinant of **A** vanishes. It can be shown that the numerator of the determinant of matrix **A** contains three radicals and is dependant of the working mode. If we manipulate properly this expression and raise it to square three times, we can obtain a polynomial of degree 16 in $x$ and $y$ (Bonev et al. 2003). This polynomial will cover all working modes. Note, however, that if $L_i = 0$, the numerator becomes a quadratic polynomial in $x$ and $y$ and that the denominator of this expression is equal to $\rho_1\,\rho_2\,\rho_3$. Unfortunately, the study of this determinant cannot characterize the motion gained by the mobile platform at Type 2 singularities.

In a Type 2 singularity, the lines normal to the directions of the prismatic joints and passing through points $C_i$ are concurrent or parallel (Fig. 6) (Bonev et al. 2003). These lines coincide with the direction of the forces $\mathbf{R}_i$ applied to the platform by the actuators.

However, we need more information for characterizing the complete kinematic behaviour of the robot inside such a singular configuration. This can be found by studying the degeneracy of the direct kinematic model. Thus, there are Type 2 singularities if:

- $\mathcal{E}(C_3)$ is an ellipse tangent to $\mathcal{L}_3$: in such a case, the directions of the three forces $\mathbf{R}_i$ intersect in one point, $W$, and the robot gains one infinitesimal rotation about this point (Fig. 6a);
- $\mathcal{L}_1$, $\mathcal{L}_2$ and $\mathcal{L}_3$ are parallel and $\mathcal{E}(C_3)$ degenerates to two lines parallel to $\mathcal{L}_1$ and $\mathcal{L}_2$ (and $\mathcal{L}_3$): in such a case, the directions of the three forces $\mathbf{R}_i$ are parallel and the robot gains one self-motion of translation (Fig. 6b);
- $\mathcal{E}(C_3)$ degenerates to a doubly-traced line segment parallel to $\mathcal{L}_3$: this case will be discussed in detail in Section 3.

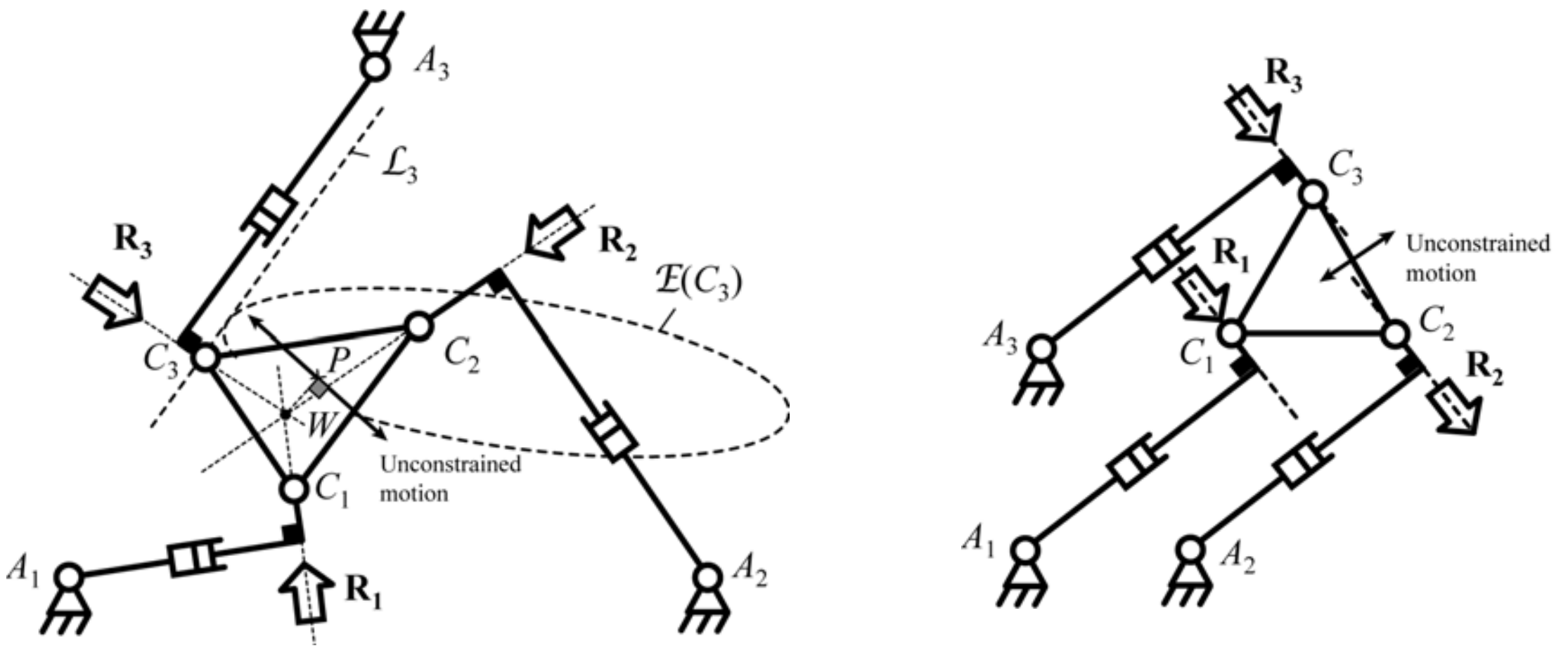

**(a) Infinitesimal rotation about *W*** **(b) Finite translation (self-motion) along the direction of the prismatic joints**

**Figure 6. Type 2 singularities of the parallel robot.**

## 3. Analysis of Self-motions

Self-motions are certainly the worst type of singularity a parallel robot can encounter. If the robot enters such a singularity, since there are infinitely many possible poses for the same active-joint variables, the information on the pose of the platform is lost. For the robot under study, one could think that such singularities exist only when $\mathcal{L}_1$, $\mathcal{L}_2$ and $\mathcal{L}_3$ are parallel. In this case, we observe the apparition of a self-motion of translation, corresponding to the case shown in Fig. 6b.

It turns out that a second more complicated case of self-motion appears when $\mathcal{E}(C_3)$ degenerates into a doubly traced line segment parallel to $\mathcal{L}_3$. This case corresponds to a Cardanic self-motion (Fig. 7).

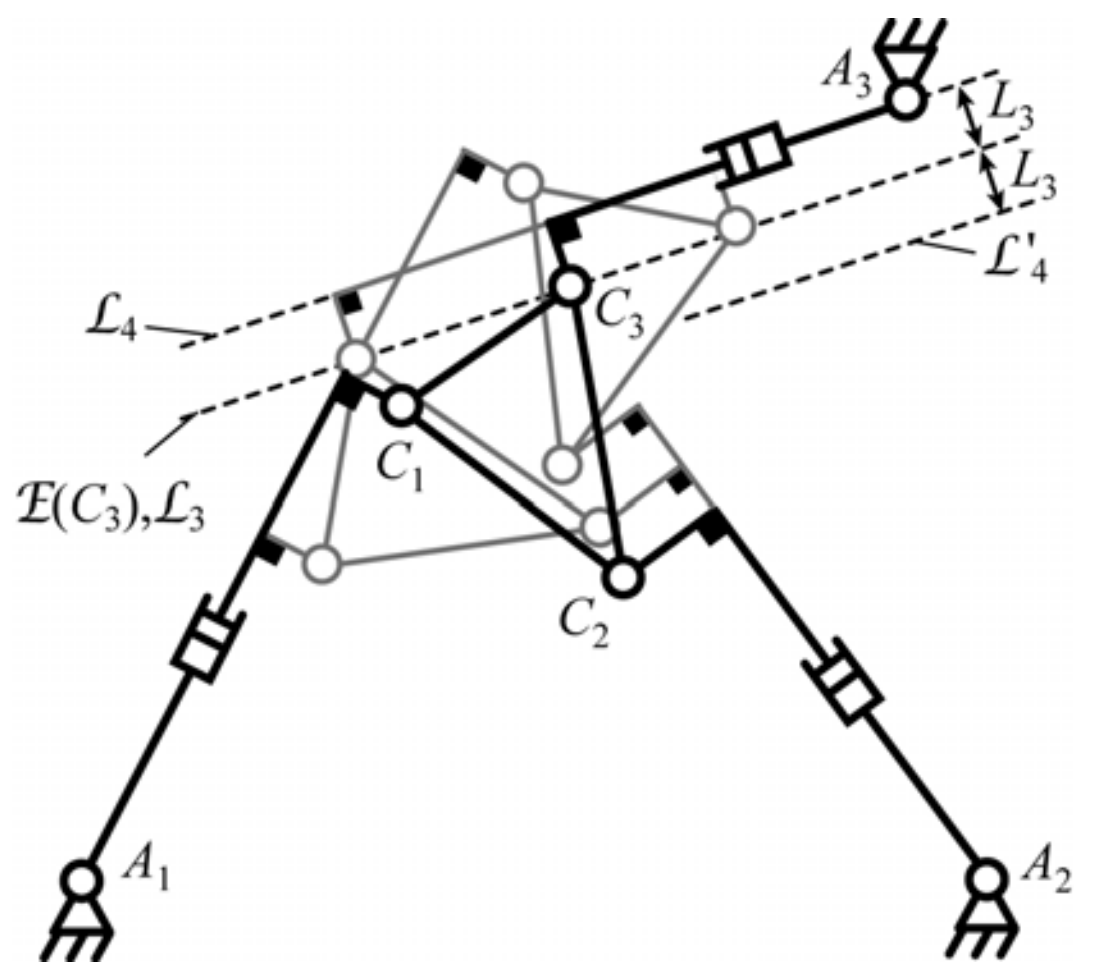

**Figure 7. Cardanic self-motion.**

Note that such a singularity is a particular case of singular configuration where the three forces $\mathbf{R}_i$ intersect at one point $W$ (Fig. 6a).

### 3.1. Design conditions leading to Cardanic self-motions

We have to find the geometric conditions which lead to Cardanic self-motions, i.e., when the ellipse $\mathcal{E}(C_3)$ degenerates into a doubly-traced line segment. This happens when $y_{C3}$ is linearly dependant on $x_{C3}$ for $\sin(\theta_1 - \theta_2) \neq 0$. Rewriting equation (13), one can obtain:

$$\mathbf{OC}_3 = \begin{bmatrix} x_{C3} \\ y_{C3} \end{bmatrix} = \begin{bmatrix} b_{11} \\ b_{21} \end{bmatrix} + \mathbf{b} \begin{bmatrix} \cos\phi \\ \sin\phi \end{bmatrix}, \text{ where } \mathbf{b} = \begin{bmatrix} b_{12} & b_{13} \\ b_{22} & b_{23} \end{bmatrix}. \tag{20}$$

$\mathcal{E}(C_3)$ will degenerate to a doubly-traced line if the determinant of matrix **b** vanishes. This would be the case if:

$$\theta_1 = \theta_2 + \varepsilon_p, \text{ where } \varepsilon_p = \alpha_p \pm \pi/2. \tag{21a}$$

As pointed out by one of the anonymous reviewers of this paper, this simple condition can also be directly obtained using the geometric properties of Cardanic motion: at each moment the intersection point between lines $\mathcal{L}_1$ and $\mathcal{L}_2$ lies on the circumcircle of the mobile platform.

Thus, for such a condition, it is possible to find through algebraic manipulations that:

$$y_{C3} = m(x_{C3} - b_{11}) + b_{21} \text{ and } \theta_3 = \theta_2 + \delta_p \tag{21b}$$

where $m = \tan\theta_3$ and $\delta_p = \beta_p/2 + n\pi$ ($n$ = 0, 1, 2, …). Once again, this condition can also be obtained using the fact that at each moment the intersection point between lines $\mathcal{L}_2$ and $\mathcal{L}_3$ lies on the circumcircle of the mobile platform. It can also be shown that lines $\mathcal{L}_1$, $\mathcal{L}_2$ and $\mathcal{L}_3$ are concurrent.

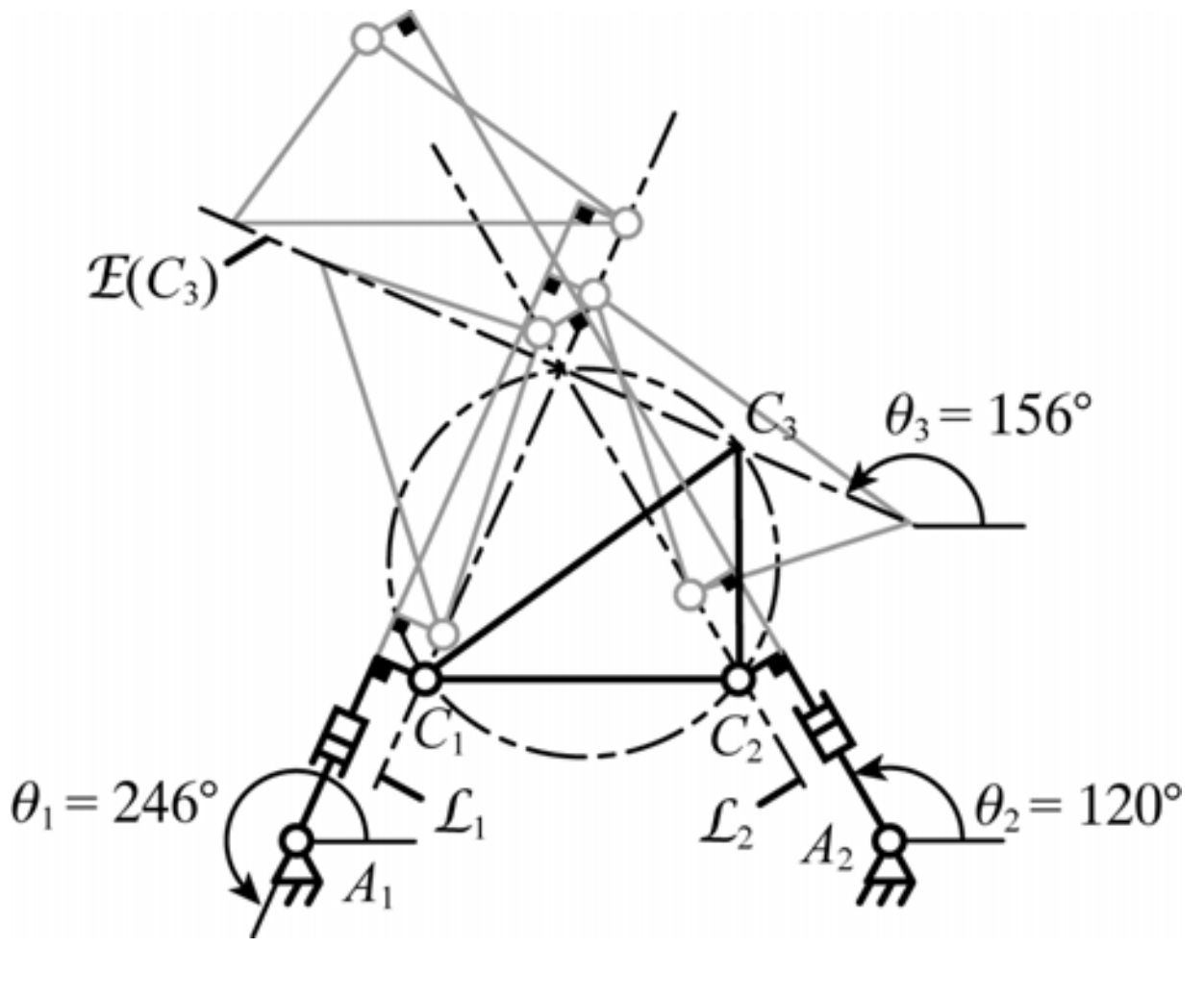

**Figure 8. Example of Cardanic motion for a 3-<u>*R*</u>*PR* planar parallel robot with $R_p$ = 0.2 m, $R_b$ = 0.35 m, $L_1 = L_2$ = 0.05 m ($L_3$ can be arbitrary), $\alpha_p$ = 36° and $\beta_p$ = 72°.**

Therefore, when $\mathcal{L}_1$ and $\mathcal{L}_2$ make an angle of $\varepsilon_p$ and $\mathcal{L}_2$ and $\mathcal{L}_3$ make an angle of $\delta_p$, the robot gains a Cardanic self-motion (Fig. 8). However, at this stage, it is not clear whether any design allows self-motions or only particular (symmetric) ones.

Let us now find the conditions for the existence of Cardanic self-motions. Introducing conditions (21a) and (21b) into equation (16), it turns out that terms $c_2$ and $c_3$ are equal to zero, therefore the equation can be simplified as:

$$c_1 = 0 . \tag{22}$$

Developing equation (22) and expressing $c_1$ as function of the sine and cosine of $\theta_2$, one obtains:

$$c_1 = d_1 \cos\theta_2 + d_2 \sin\theta_2 + d_3 = 0 , \tag{23}$$

where

$$d_1 = R_b \left(\sin(\delta_p + \alpha_b - \beta_b) - \sin(\alpha_b - \delta_p)\right), \tag{24}$$

$$\begin{aligned} d_2 = {} & R_b \left(\sin(\delta_p + \varepsilon_p + \alpha_b - \beta_b) - \sin(\delta_p - \varepsilon_p + \alpha_b - \beta_b) - \sin(\varepsilon_p + \alpha_b - \delta_p)\right) / \sin\varepsilon_p \\ & - R_b \left(\sin(-\varepsilon_p + \alpha_b - \delta_p) + 2\sin(-\varepsilon_p + \alpha_b + \delta_p)\right) / \sin\varepsilon_p \end{aligned} , \tag{25}$$

$$d_3 = \frac{L_1 \sin\delta_p - L_2 \sin(\delta_p - \varepsilon_p) - L_3 \sin\varepsilon_p}{\sin\varepsilon_p} . \tag{26}$$

Thus, two cases for the cancellation of equation (23) must be examined:

(a) when equation (23) is satisfied only for some sets of active-joint angles;

(b) when equation (23) is satisfied for any $\theta_2$, , which is only possible if $d_1 = d_2 = d_3 = 0$.

Let us begin with the first case. The sets of active-joint variables satisfying equation (23) can be found as:

$$\theta_{2p} = 2\tan^{-1}\left( \frac{-d_2 + \sqrt{d_2^2 + d_1^2 - d_3^2}}{d_3 - d_1} \right), \tag{27a}$$

$$\theta_{2m} = 2\tan^{-1}\left(\frac{-d_2 - \sqrt{d_2^2 + d_1^2 - d_3^2}}{d_3 - d_1}\right). \tag{27b}$$

As angles $\delta_p$ and $\varepsilon_p$ are defined with $n\pi\,(n = 0, 1, 2,\ldots)$, the maximal number of sets of active-joint variables is equal to eight, depending of the working modes. As pointed out by one of the anonymous reviewers, these solutions correspond to the intersection of six limaçons defined as the loci of the intersection point between lines $\mathcal{L}_1$, $\mathcal{L}_2$ and $\mathcal{L}_3$ (which are concurrent for a Cardanic self-motion) when varying angle $\theta_2$. Thus, the robot can have Cardanic self-motions for a maximum of eight sets of (or infinitely many) active-joint angles.

Now, the more useful result is that there obviously exist designs without Cardanic self-motions. The condition for non-existence of Cardanic self-motions is simply the condition that prevents equation (23) to have real solutions, i.e.:

$$d_3^2 > d_2^2 + d_1^2. \tag{28}$$

Considering the simple case where the base and platform are similar (or even equilateral) triangles and the offsets are equal $L = L_1 = L_2 = L_3$, and introducing these new parameters in equation (28), it can be found that the condition of non-existence of Cardanic self-motions is:

$$L \neq 0. \tag{29}$$

Thus, there exist simple symmetric designs without Cardanic self-motions.

Now, we saw that Cardanic self-motions appear (or not) for only several active-joint sets, whereas it is possible to see in (Chablat et al. 2006), for a particular design of 3-*R̲PR* planar parallel robot with congruent equilateral base and platform triangles, that if condition (21) is satisfied, there exists an infinity of active-joint sets for which the robot gains a Cardanic self-motion. Thus, there must be design conditions for the robot to have Cardanic self-motion for any value of angle $\theta_2$.

The second possibility for cancelling equation (23) consists of the cancellation of terms $d_i$ of equations (24) to (26). Resolving these three equations leads to:

$$L_1 \sin\delta_p - L_2 \sin(\delta_p - \varepsilon_p) - L_3 \sin\varepsilon_p = 0 \quad (30)$$

and

$$\alpha_b = \alpha_p \text{ and } \beta_b = \beta_p. \quad (31)$$

Thus, the base and the mobile platform should be similar triangles and condition (30) on the offsets must hold. Such conditions for Cardanic self-motions do not depend of the value of angle $\theta_2$, as previously demonstrated in (Chablat et al. 2006).

In summary, Cardanic self-motions can be avoided by well constraining the design parameters of the 3-*R̲PR* planar parallel robot (equation 28). In the worst case, if the base and the mobile platform are similar and if $L_1 \sin\delta_p - L_2 \sin(\delta_p - \varepsilon_p) - L_3 \sin\varepsilon_p = 0$, there are Cardanic self-motions for infinitely many active-joint sets. Finally, if one wants to have similar or even equilateral base and platform triangles, one way of completely avoiding self-motions is to use equal non-zero offsets.

3.2. Kinematic analysis of the Cardanic self-motion

Let us now analyse the allowable displacement of the centre *P* of the platform when the base and the mobile platform are similar triangles, $\theta_1 = \theta_2 + \varepsilon_p$, $\theta_3 = \theta_2 + \delta_p$, $L_1 \sin\delta_p - L_2 \sin(\delta_p - \varepsilon_p) - L_3 \sin\varepsilon_p = 0$. The expressions of the coordinates of point *P,* function of $\theta_2$, are found using the following closure equation:

$$\mathbf{OP} = \mathbf{OA_2} + \mathbf{A_2B_2} + \mathbf{B_2C_2} + \mathbf{C_2P}. \quad (32)$$

Developing this expression, one can obtain:

$$\mathbf{OP} = \begin{bmatrix} x \\ y \end{bmatrix} = \begin{bmatrix} x_{A2} \\ y_{A2} \end{bmatrix} + \rho_2 \begin{bmatrix} \cos\theta_2 \\ \sin\theta_2 \end{bmatrix} + L_2 \begin{bmatrix} -\sin\theta_2 \\ \cos\theta_2 \end{bmatrix} - R_p \begin{bmatrix} \cos(-\alpha_p + \phi) \\ \sin(-\alpha_p + \phi) \end{bmatrix}, \tag{33}$$

where the expression of $\rho_2$ is given by equation (12). Developing and introducing equations (21), (30) and (31) in (33), it can be found that:

$$\mathbf{OP} = \begin{bmatrix} R_p \cos(\alpha_p + 2\theta_2 - \phi) - R_b \cos(\alpha_p + 2\theta_2) - L_2 \sin(\theta_2 + \alpha_p) - L_1 \cos\theta_2 \\ R_p \sin(\alpha_p + 2\theta_2 - \phi) - R_b \sin(\alpha_p + 2\theta_2) + L_2 \cos(\theta_2 + \alpha_p) - L_1 \sin\theta_2 \end{bmatrix} \tag{34}$$

From the previous expression, it is possible to conclude that, in such a particular configuration, varying the orientation $\phi$ of the mobile platform, point $P$ moves on a circle $S$ centred in $O$' whose radius is $R_p$ (Fig. 9). The coordinates of point $O$' are defined by:

$$\mathbf{OO'} = -R_b \begin{bmatrix} \cos(\alpha_p + 2\theta_2) \\ \sin(\alpha_p + 2\theta_2) \end{bmatrix} + L_2 \begin{bmatrix} -\sin(\theta_2 + \alpha_p) \\ \cos(\theta_2 + \alpha_p) \end{bmatrix} - L_1 \begin{bmatrix} \cos\theta_2 \\ \sin\theta_2 \end{bmatrix}. \tag{35}$$

Computing the expressions of the coordinates of point $W$, the intersection point of the three wrenches $\mathbf{R}_i$, one obtains:

$$\mathbf{OW} = \begin{bmatrix} 2R_p \cos(\alpha_p + 2\theta_2 - \phi) - R_b \cos(\alpha_p + 2\theta_2) - L_2 \sin(\theta_2 + \alpha_p) - L_1 \cos\theta_2 \\ 2R_p \sin(\alpha_p + 2\theta_2 - \phi) - R_b \sin(\alpha_p + 2\theta_2) + L_2 \cos(\theta_2 + \alpha_p) - L_1 \sin\theta_2 \end{bmatrix}. \tag{36}$$

Thus, $W$ is located on a circle $\mathcal{K}$ centred in $O$' whose radius is $2R_p$. It is also possible to observe that the platform and vector **O'P** rotate in opposite senses.

One can rewrite expression (34) as follows:

$$\mathbf{OP} = \begin{bmatrix} R \cos(\eta + 2\theta_2) - L_2 \sin(\theta_2 + \alpha_p) - L_1 \cos\theta_2 \\ R \sin(\eta + 2\theta_2) + L_2 \cos(\theta_2 + \alpha_p) - L_1 \sin\theta_2 \end{bmatrix} \tag{37}$$

with

$$R = \sqrt{R_b^2 + R_p^2 - 2R_b R_p \cos\phi} \text{ and } \eta = \tan^{-1}\left(-\frac{R_p \sin(\phi - \alpha_p) - R_b \sin\alpha_p}{R_p \cos(\phi - \alpha_p) - R_b \cos\alpha_p}\right). \quad (38)$$

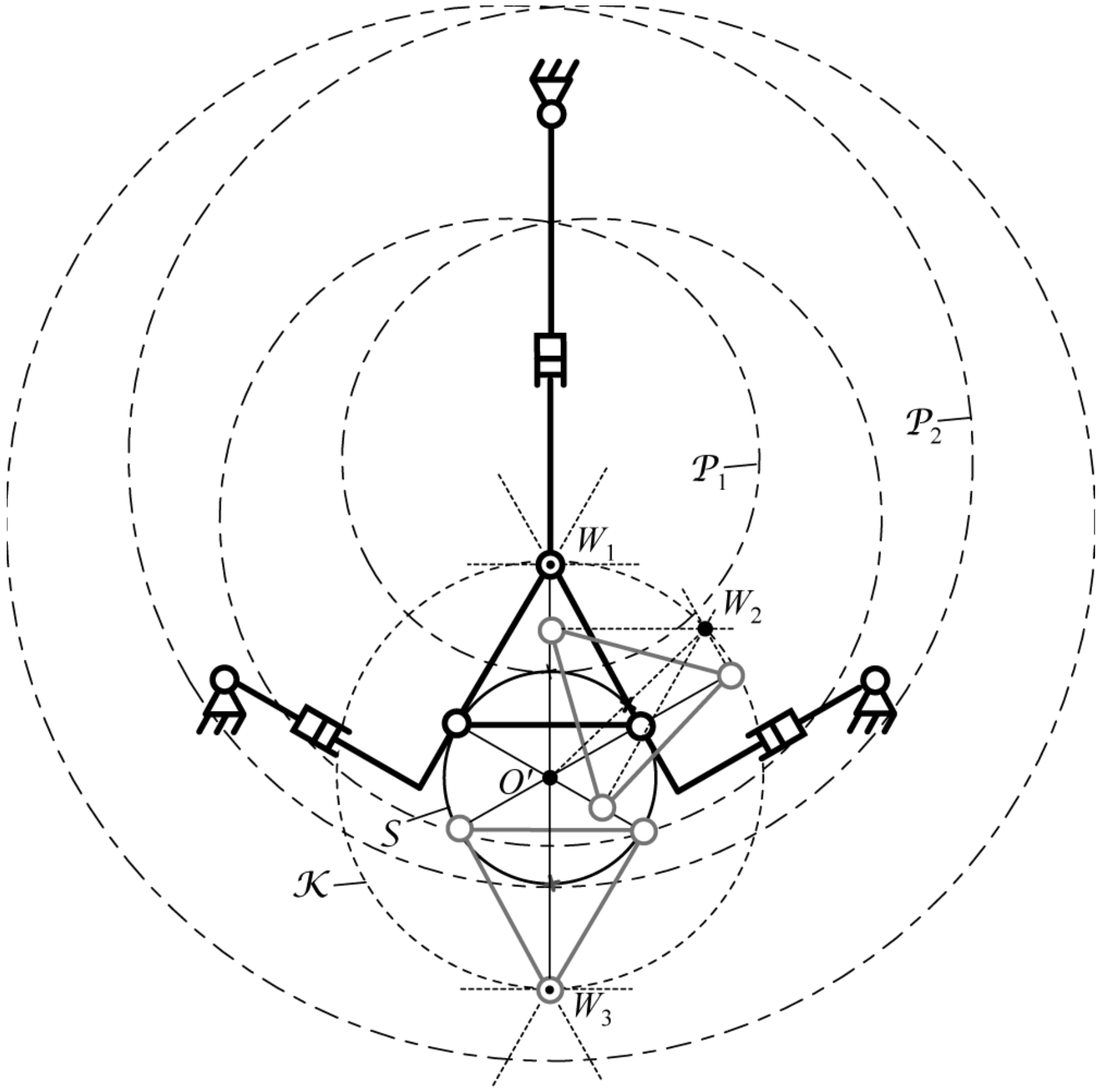

**Figure 9. Schematics of a Cardanic self-motion of a robot with $R_p$ = 0.1 m, $R_b$ = 0.35 m, $L_1 = L_2$ = 0.07 m, $L_3$ = 0 m, $\alpha_b$ = 30° and $\beta_b$ = 120°.**

For a given angle $\phi$ and active-joint angle $\theta_2$, equation (37) represents the singularity loci (for the Cardanic self-motions only) of the robot with specified parameters. The obtained result corresponds to the parametric expression of an epicycloid $\mathcal{P}$. The epicycloids $\mathcal{P}_1$ and $\mathcal{P}_2$ represented in Fig. 9 are the curves corresponding to angles $\phi = 0$ and $\phi = \pi$ respectively.

## 4. Example and experimental validations

A prototype of a new decoupled 4-DOF parallel robot called PAMINSA (Parallel Robot of the INSA, Fig. 10) was developed in INSA de Rennes (Briot and Arakelian 2007). Such a robot with Schoenflies motions allows the decoupling of the displacements in a horizontal plane (two translations along the $\boldsymbol{x}$ and $\boldsymbol{y}$ axes and one rotation about axes parallel to $\boldsymbol{z}$) from the translation along a vertical axis (for details, see (Briot and Arakelian 2007)). Thus, this decoupling allows the separation of the control laws between two different models:

- a model for the horizontal displacements equivalent to the control model of the 3-$\underline{R}PR$ planar parallel robot (Fig. 11a);
- a linear model for the vertical translation due to the use of the pantograph linkage (Fig. 11b).

Thus, PAMINSA presents the same Type 2 singularities as a symmetric 3-$\underline{R}PR$ planar parallel robot, which will be studied in this section. Indeed, the planar projection of the prototype of the PAMINSA robot corresponds to a 3-$\underline{R}PR$ planar parallel robot whose base and platform are non-identical equilateral triangles and whose offsets are zero, $L_i = 0$. These conditions correspond to a robot with infinitely many Cardanic self-motions within its workspace.

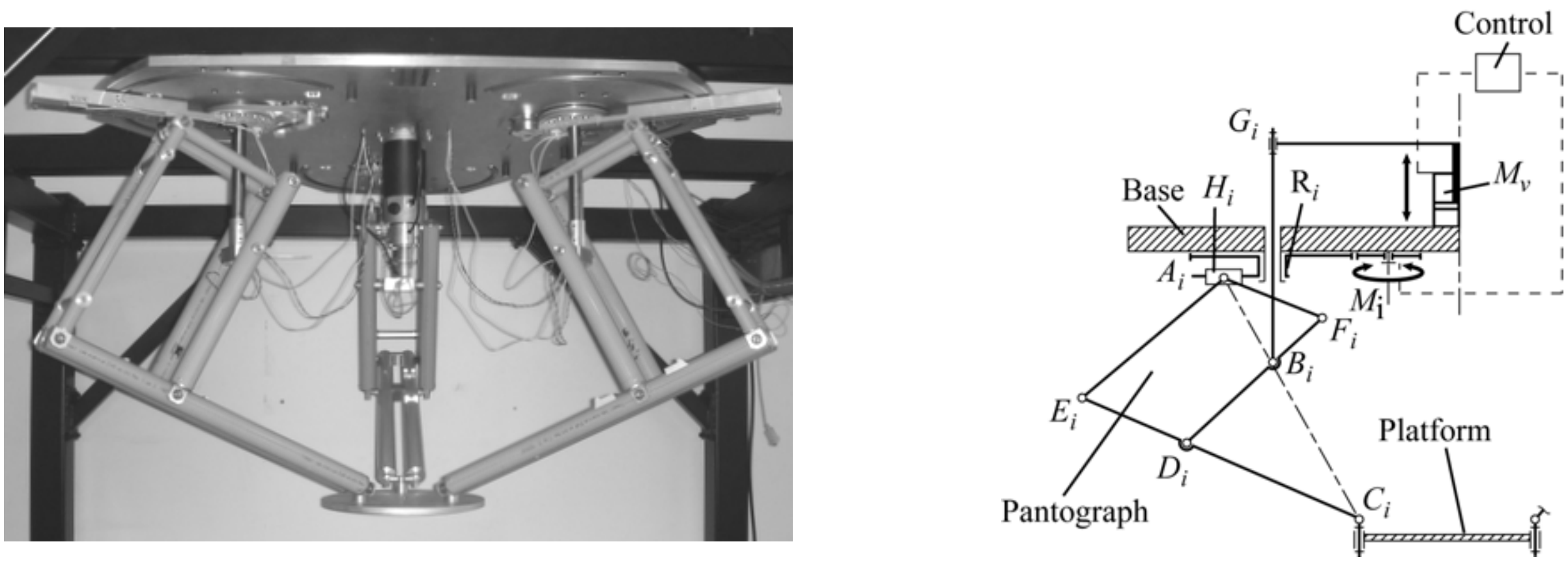

**(a) Prototype of the PAMINSA robot** **(b) kinematic chain**

**Figure 10. The PAMINSA parallel robot.**

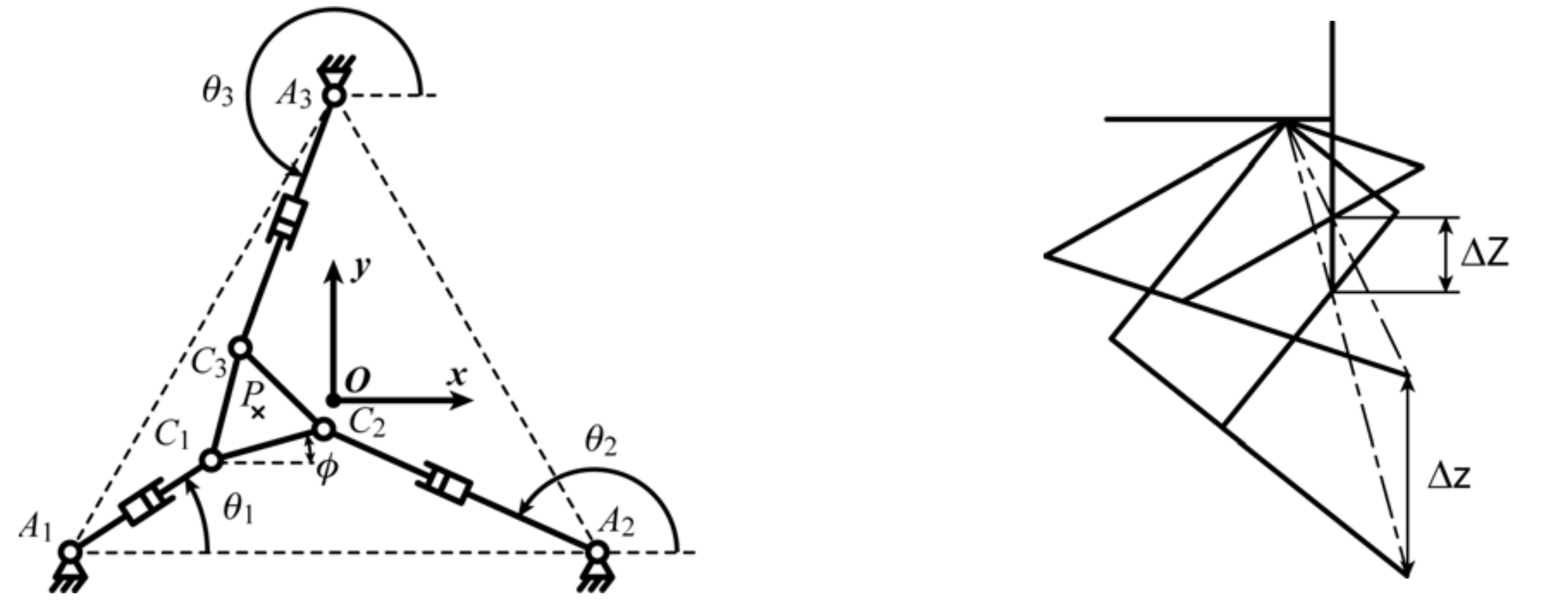

**(a) model for the planar displacements** **(b) model for the vertical translations**

**Figure 11. The control models for the PAMINSA parallel robot.**

Introducing these constraints in matrix **A** of equation (3), we can find the determinant of this matrix as

$$D = \frac{2R_p \cos\alpha_p (\sin(\alpha_p - \beta_p) - \sin\alpha_p)}{\rho_1\rho_2\rho_3}(R_b \cos\phi - R_p)(x^2 + y^2 - (R_b^2 + R_p^2 - 2R_pR_b\cos\phi)) \,. \qquad (39)$$

Type 2 singularity loci for the PAMINSA occur when the above expression vanishes (Briot and Arakelian 2007). Thus, the robot is in a Type 2 singularity when:

$$\rho_i = \pm\infty \text{, for } i = 1, 2 \text{ or } 3 \tag{40}$$

or

$$\phi = \phi_s = \pm\cos^{-1}(R_p / R_b) \tag{41}$$

or

$$x^2 + y^2 = R_b^2 + R_p^2 - 2R_b R_p \cos\phi \tag{42}$$

Condition (40) implies that the platform is located at an infinite distance from the centre of the base frame. This is equivalent to the fact that the three legs of the robot are parallel (Fig. 6b). Condition (41) implies that the robot gains one degree of freedom for any position ($x$, $y$) of the workspace, for a fixed platform angle $\phi_s$. Finally, condition (42) implies that the robot gains one degree of freedom when point $P$ is located on a circle centred at $O$ whose radius is $R = \sqrt{R_b^2 + R_p^2 - 2R_b R_p \cos\phi}$. Thus, we have to find which of the last two conditions correspond to Cardanic self-motions.

Introducing the constraints $L_i = 0$, $\alpha_b = \alpha_p$ and $\beta_b = \beta_p$ into equation (34), one can find:

$$\mathbf{OP} = \begin{bmatrix} x \\ y \end{bmatrix} = \begin{bmatrix} R_p \cos(\alpha_p + 2\theta_2 - \phi) - R_b \cos(\alpha_p + 2\theta_2) \\ R_p \sin(\alpha_p + 2\theta_2 - \phi) - R_b \sin(\alpha_p + 2\theta_2) \end{bmatrix}. \tag{43}$$

Raising the norm of vector **OP** to square, we obtain equation (43). Thus, this particular design of 3-<u>*R*</u>*PR* planar parallel robot gains one Cardanic self-motion when the end effector is positioned on a circle $\mathcal{P}$ centred at $O$ and with radius equal to $R = \sqrt{R_b^2 + R_p^2 - 2R_b R_p \cos\phi}$ (Fig. 9). The circles $\mathcal{P}_1$ and $\mathcal{P}_2$ represented in Fig. 12 are the circles $\mathcal{P}$ corresponding to angles $\phi = 0$ and $\phi = \pi$, respectively .

Note that, for the angle $\phi_s$, the robot gains one infinitesimal degree of freedom at any position, except if point $P$ is located on a circle centred in $O$ whose radius is equal to $R_s = \sqrt{R_b^2 + R_p^2 - 2R_b R_p \cos\phi_s}$ . Such position still corresponds to a Cardanic self-motion. Moreover, for $R_p$=$R_b$, the angle $\phi_s$ corresponds to a self-motion of translation (Chablat et al. 2006). This means that, when the platform centre is located on the circle $\mathcal{P}_1$, the platform gains two self-motions at the same time.

Observing equation (43), it is possible to conclude that the degree of freedom gained is motion along a circle $\mathcal{S}$ centred in $O$' whose radius is $R_p$. The coordinates of point $O$' are:

$$\mathbf{OO'} = -R_b \begin{bmatrix} \cos(\alpha_p + 2\theta_2) \\ \sin(\alpha_p + 2\theta_2) \end{bmatrix}. \tag{44}$$

Note that the circle $\mathcal{S}$ is tangent to circles $\mathcal{P}_1$ and $\mathcal{P}_2$. This means that the maximal singularity-free workspace is delimited by the circle $\mathcal{P}_1$. The radius of the circle $\mathcal{P}_1$ is equal to:

$$R_1 = \left| R_b - R_p \right|. \tag{45}$$

Dividing equation (45) by $R_b$ yields

$$\nu = R_1 / R_b = \left| 1 - R_p / R_b \right|. \tag{46}$$

Thus, the smaller the ratio $R_p/R_b$, the greater the value of $\nu$. So it is possible conclude that, for having a larger singularity-free workspace, the rate $R_p/R_b$ has to be smaller. However, the smaller the mobile platform with respect to the base, the less accurate is its orientation.

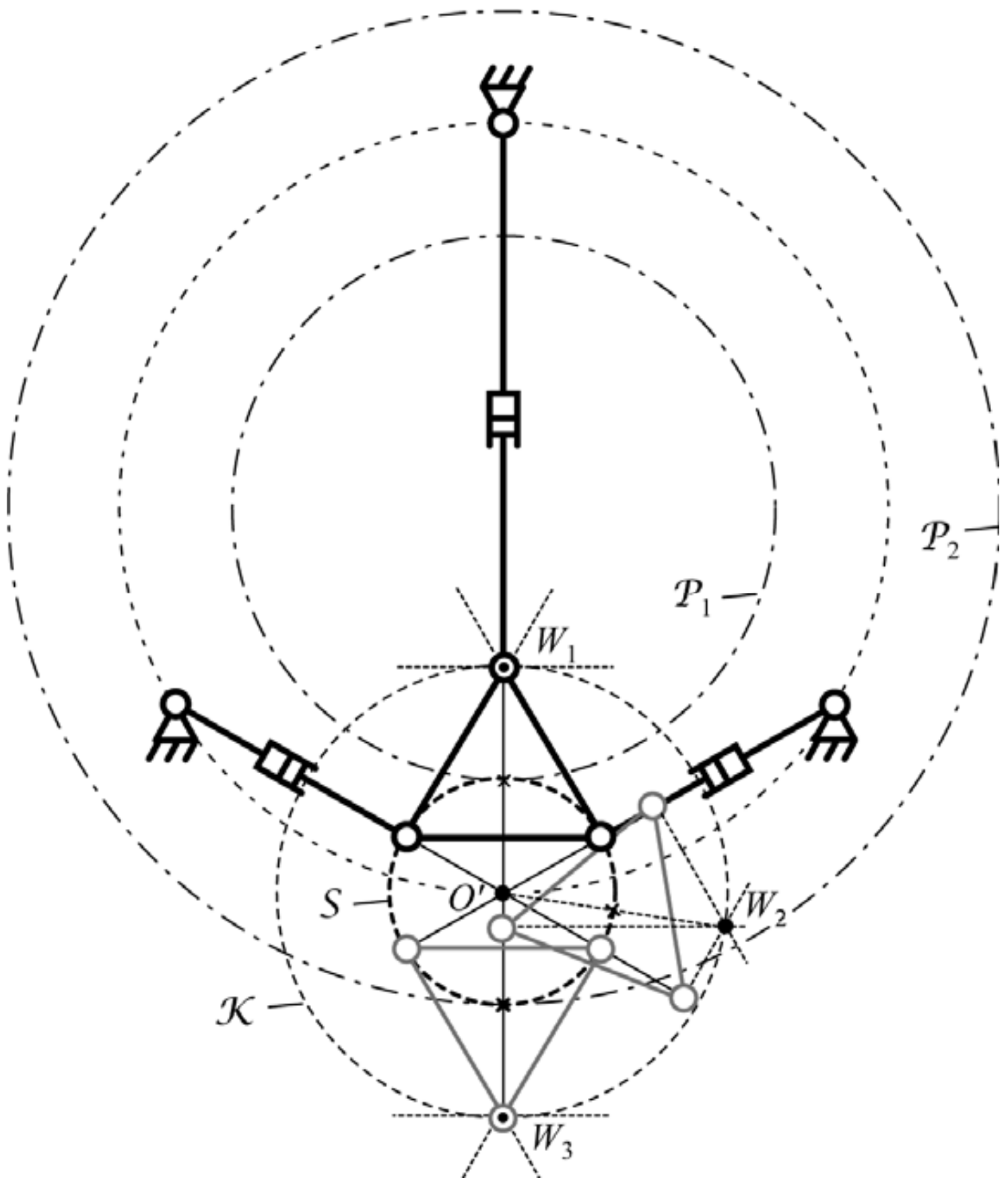

**Figure 12. Schematics of a Cardanic self-motion for a robot with $R_p$ = 0.1 m, $R_b$ = 0.35 m, $\alpha_b$ = 30° and $\beta_b$ = 120°.**

In order to demonstrate the previous results, we have positioned the PAMINSA prototype in a singular configuration with Cardanic self-motion ($x$ = 0 m, $y$ = −0.25 m, $\phi$ = 0°). This position is shown on Fig. 13(g). For such a configuration, the three actuators are blocked. However, it is possible to see on Figs. 13(a) to 13(e) that the platform is not constrained and undergoes a Cardanic self-motion when external force is applied to the platform.

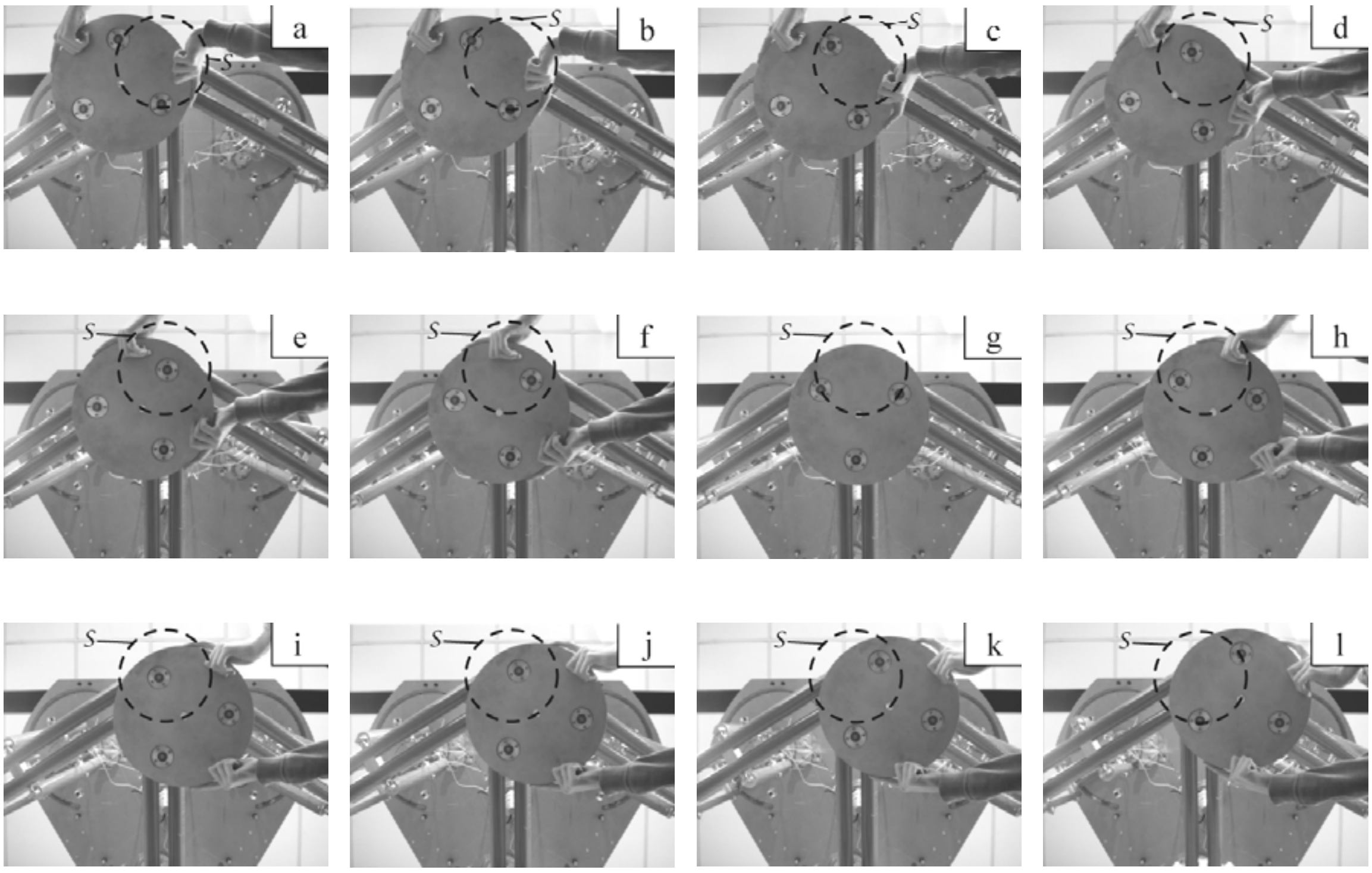

**Figure 13. Cardanic self-motion of the mobile platform of the PAMINSA prototype starting from the configuration $x$ = 0 m, $y$ = -0.25 m, $\phi$ = 0°.**

## 5. Conclusions

In this paper, the singular configurations of general 3-*<u>R</u>PR* planar parallel robots were studied. It was shown that a general 3-*<u>R</u>PR* robot can have Cardanic self-motions for none, up to eight, or infinitely many active-joint sets. The conditions for having no self-motions or having self-motions for infinitely many active-joint sets were explicitly derived. It was shown, for example, that designs with similar (or even equilateral) base and platform triangles and equal offsets have no self-motions as long as the offsets non-zero.

## 6. Acknowledgements

The authors would like to thank the anonymous reviewers for their invaluable comments.

## Appendix

Expressions of $a_{ji}$ ($j$ = 1, 2, i=1, 2, 3):

$$a_{11} = \frac{(x_{A2} - x_{A1})\sin\theta_2 + L_1\cos(\theta_2 - \theta_1) - L_2}{\sin(\theta_2 - \theta_1)} \quad a_{12} = -\frac{2R_p \sin\theta_2 \cos\alpha_p}{\sin(\theta_2 - \theta_1)} \quad a_{13} = \frac{2R_p \cos\theta_2 \cos\alpha_p}{\sin(\theta_2 - \theta_1)}$$

$$a_{21} = \frac{(x_{A2} - x_{A1})\sin\theta_1 - L_2\cos(\theta_2 - \theta_1) + L_1}{\sin(\theta_2 - \theta_1)} \quad a_{22} = -\frac{2R_p \sin\theta_1 \cos\alpha_p}{\sin(\theta_2 - \theta_1)} \quad a_{23} = \frac{2R_p \cos\theta_1 \cos\alpha_p}{\sin(\theta_2 - \theta_1)}$$

Expressions of $b_{ji}$ ($j$ = 1, 2, i=1, 2, 3):

$$b_{11} = x_{A1} + a_{11}\cos\theta_1 - L_1\sin\theta_1 \qquad b_{21} = y_{A1} + a_{11}\sin\theta_1 + L_1\cos\theta_1$$

$$b_{12} = a_{12}\cos\theta_1 + 2R_p\cos\left(\alpha_p - \frac{\beta_p}{2}\right)\cos\frac{\beta_p}{2} \qquad b_{22} = a_{12}\sin\theta_1 + 2R_p\cos\left(\alpha_p - \frac{\beta_p}{2}\right)\sin\frac{\beta_p}{2}$$

$$b_{13} = a_{13}\cos\theta_1 - 2R_p\cos\left(\alpha_p - \frac{\beta_p}{2}\right)\sin\frac{\beta_p}{2} \qquad b_{23} = a_{13}\sin\theta_1 + 2R_p\cos\left(\alpha_p - \frac{\beta_p}{2}\right)\cos\frac{\beta_p}{2}$$

Expressions of $c_i$ (i=1, 2, 3):

$$c_1 = (b_{21} - y_{A3})\cos\theta_3 + (x_{A3} - b_{11})\sin\theta_3 - L_3, \; c_2 = b_{22}\cos\theta_3 - b_{12}\sin\theta_3, \; c_3 = b_{23}\cos\theta_3 - b_{13}\sin\theta_3.$$

Figure Captions

Figure 1. Schematic representation of the 3-*RPR* planar parallel robot under study.

Figure 2. Parameterisation of the base and platform triangles.

Figure 3. The two inverse kinematic solutions of the $i^{th}$ leg of the robot.

Figure 4. Type 1 singularity.

Figure 5. Geometric interpretation of the direct kinematics.

Figure 6. Type 2 singularities of the parallel robot.

Figure 7. Cardanic self-motion.

Figure 8. Example of Cardanic motion for a 3-*RPR* planar parallel robot with $R_p = 0.2$ m, $R_b = 0.35$ m, $L_1 = L_2 = 0.05$ m ($L_3$ can be arbitrary), $\alpha_p = 36°$ and $\beta_p = 72°$..

Figure 9. . Schematics of a Cardanic self-motion of a robot with $R_p = 0.1$ m, $R_b = 0.35$ m, $L_1 = L_2 = 0.07$ m, $L_3 = 0$ m, $\alpha_b = 30°$ and $\beta_b = 120°$.

Figure 10. The PAMINSA parallel robot.

Figure 11. The control models for the PAMINSA parallel robot.

Figure 12. Schematics of a Cardanic self-motion for a robot with $R_p = 0.1$ m, $R_b = 0.35$ m, $\alpha_b = 30°$ and $\beta_b = 120°$.

Figure 13. Cardanic self-motion of the mobile platform of the PAMINSA prototype starting from the configuration $x = 0$ m, $y = -0.25$ m, $\phi = 0°$.